# Bringing Cartoons to Life: Towards Improved Cartoon Face Detection and Recognition Systems

Saurav Jha
mail@sauravjha.com.np
MNNIT Allahabad, India

Nikhil Agarwal
nikil.agar@gmail.com
MNNIT Allahabad, India

Suneeta Agarwal
suneeta@mnnit.ac.in
MNNIT Allahabad, India

**ABSTRACT**

Given the recent deep learning advancements in face detection and recognition techniques for human faces, this paper answers the question "how well would they work for cartoons'?" - a domain that remains largely unexplored until recently, mainly due to the unavailability of large scale datasets and the failure of traditional methods on these. Our work studies and extends multiple frameworks for the aforementioned tasks. For face detection, we incorporate the Multi-task Cascaded Convolutional Network (MTCNN) architecture [1] and contrast it with conventional methods. For face recognition, our two-fold contributions include: (i) an inductive transfer learning approach combining the feature learning capability of the Inception v3 network [2] and the feature recognizing capability of Support Vector Machines (SVMs), (ii) a proposed Hybrid Convolutional Neural Network (HCNN) framework trained over a fusion of pixel values and 15 manually located facial keypoints. All the methods are evaluated on the Cartoon Faces in the Wild (IIIT-CFW) database [3]. We demonstrate that the HCNN model offers stability superior to that of Inception+SVM over larger input variations, and explore the plausible architectural principles. We show that the Inception+SVM model establishes a state-of-the-art (SOTA) F1 score on the task of gender recognition of cartoon faces. Further, we introduce a small database hosting location coordinates of 15 points on the cartoon faces belonging to 50 public figures of the IIIT-CFW database.

## 1 INTRODUCTION

The exponential rise in digital media over the recent years has in turn elevated the trend of cartoonifying subjects, as they serve to inhabit a wide range of life aspects such as providing home schools for children [4], enhancing the teaching process and academic achievement [5] as well as depicting one's opinion on the practices of society (through political cartoons and comics journalism). In contrast to the standard drawings, the subjects appearing in cartoons[1] possess features exaggerated in ways that often lead to the deviation of such faces from the implicit humanly attributes (e.g. facial symmetry violation, unnatural skin tone, anomalous facial outline, etc.) presumed by most of the benchmark detection [6–8] and recognition techniques [9–11]. While such techniques have found their wide usage for humans in day-to-day appliances such as biometric scanners and healthcare equipment, the spectacular rise in the cartoon industry has inflated the need for similar techniques for cartoon faces with some prominent applications including: (i) Incorporation in image search engines for searching the web for similar cartoons. (ii) Integration with screen readers to assist visually impaired people understand cartoon movies. (iii) Help content-control softwares censor inappropriate cartoon images on social media.

Our work targets meeting the above mentioned goal by leveraging deep learning systems that are capable of more accurately detecting and recognizing the cartoon faces along with providing stable results over greater artistic variations of the facial features. With the advent of large scale cartoon databases like the IIIT-CFW database [3] (Section 4.1), we depict that such systems can be effectively trained and evaluated over reasonably adequate samples. To the best of our knowledge, these are the first ever deep neural detection and recognition systems to be built specifically for cartoon faces. The contributions of our work are four-fold:

- We explore, incorporate and extend some of the current research on face detection and recognition into our framework for cartoon faces.

- For face detection (3.1), we exploit the MTCNN [1] framework, and contrast its performance with that of the Histogram of Oriented Gradients (HOG) features [12] and Haar features [13]. Our study further yields substantial improvements over the SOTA Jaw contour and symmetry based cartoon face detector [14], as described in Section 2.

- For face recognition (Section 4.5), we study two deep neural network frameworks: the former architecture employs a pre-trained GoogleNet Inception v3 based Convolutional Neural Network (CNN) architecture [2] as feature extractor assisted by SVM [15] and Gradient Boosting (GB) [16] classifiers as feature recognizer, whilst the latter is a proposed HCNN framework that leverages the pixel values of images along with the location coordinates of 15 facial keypoints in an end-to-end fashion. We demonstrate that both of these frameworks outperform the SOTA [14] in terms of F-measure (Section 4.5).

- We investigate the performance of the recognition models under various input constraints governed by a number of metrics for classifying characters and gender of the cartoon faces. For both the systems, we evaluate the effectiveness of the keypoints extraction mechanism (Section 4.4.2) and empirically show that the inclusion of the facial keypoints location results in a 5.87% gain on the top-5 error rate of the HCNN model (Section 4.5). Our further analysis demonstrate that while the HCNN model offers enhanced stability over the Inception v3+SVM model as the number of classes increase, the Inception+SVM model thus employed on the gender recognition task establishes a SOTA F-measure of 0.910 (Section 4.6).

## 2 RELATED WORKS

A greater amount of the previous researches on the subject of cartoon and comic characters recognition revolve around the task of **classifying image-level video into cartoon/non-cartoon genre**, dating

---

[1] By cartoons, we refer to cartoons, caricatures, and comics.

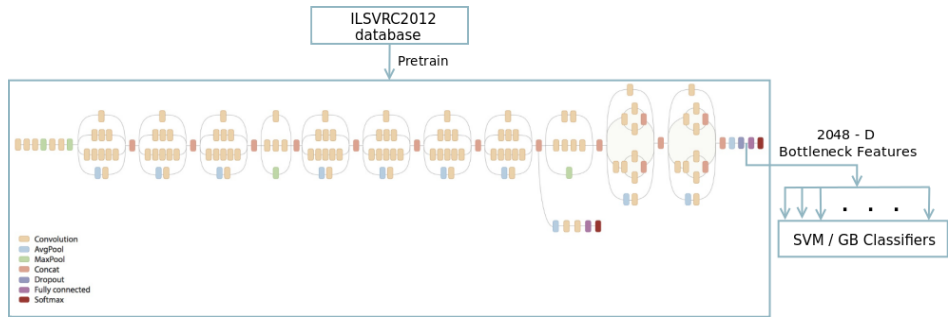

Figure 1: Inductive transfer framework for face recognition: the 2048D bottleneck features extracted by the Inception v3 model are fed into SVM and GB classifiers

back to the work of Glasberg *et al.* [17] classifying 100 MPEG-2 video sequences by combining the visual features through a multilayered perceptron followed by a fusion of the concerned audio features extracted from consecutive frames. Following them, Glasberg *et al.* [18] used a probability-based approach employing two Hidden Markov Models and five visual descriptors (features) to achieve an improved performance. Ionescu *et al.* [19] used temporal and color based content descriptors to achieve image-level classification of animated contents over 159 hours of video footage. However, the task of cartoon face detection and recognition accounts for more diverse feature contemplation than the aforementioned tasks as the individual cartoon images possess greater artistic variations among each other compared to the artistically similar characters appearing in the consecutive video frames.

Takayama *et al.* [14] presented the first relevant work concerning **cartoon face detection and recognition**. For detection, they used jaw contour and symmetry as two criteria to evaluate whether a segmented region based upon skin color and edges is a face or not while for recognition, they extracted three features (skin color, hair color and hair quantity) from each image and distinguished the correct input image class based on the similarity of feature vectors. Their method is nonetheless limited to color images with skin color near to real people and targets mainly frontal posture. Prior to this, packages such as the AnimeFace 2009[2] would use simple perceptron architectures trained over millions of image data to judge whether the face region candidates of anime faces are actually faces or not.

**Deep learning based approaches** have only come up recently. Nguyen *et al.* [20] performed comic characters detection by applying the YOLOv2 model [21] to predict the location coordinates of the bounding boxes with respect to the location of the SxS cells grid formed over the image. The Manga FaceNet proposed by Chu and Li [22] offers a CNN based architecture for detecting Manga faces (i.e., Japanese cartoons and comics). Although these frameworks present good baseline on face detection and recognition of cartoon and comic characters, a lot of the state-of-the-art methods available today remain unexplored on the tasks.

## 3 METHODOLOGY

Section 3.1 and 3.2 describe the detection and recognition models. The cartoon database used is described in Section 4.1. The **normalized images** refer to 96x96 resized images preserving the original aspect ratio.

### 3.1 Face Detection

The MTCNN [1] architecture offers a deep cascaded multi-task framework with three sequential deep CNNs: the Proposal Net, the Residual Net and the Output Net. The input to the proposal net is an image pyramid formed by resizing the input image to different scales. Each subsequent layer then performs candidate window calibration using the estimated bounding box regression vectors, merges the highly overlapped candidates, thereby outputting a final face bounding box with five facial landmarks' positions. Compared to its precursor [23], MTCNN avails three major tweaks: reduced number of filters and smaller filter sizes (3x3) lessen the computational burden while increased depth of network improves the performance. We further employ the Haar features [13] and the HOG features [12] based detectors for securing baseline results in order to gain deeper insights into the performance of the MTCNN framework.

---
[2]https://github.com/nagadomi/animeface-2009

### 3.2 Face Recognition

We experiment on two different face recognition techniques:

#### 3.2.1 Inductive transfer using Inception v3 + SVM/GB

We employ the GoogleNet Inception v3 architecture [2] pre-trained on the ImageNet Large Scale Visual Recognition Challenge's 2012 (ILSVRC2012) database for extracting 2048 dimensional features from each normalized image. The Inception architectures are well-known for availing auxiliary classifiers (a.k.a. side-heads) in addition to the main classifier for achieving more stable learning in the latter training stages. The major tweak in the Inception v3 network (compared to its antecedent [24]) lies in factoring the first 7x7 convolutional layer into a sequence of 3x3 convolutional layers followed by a batch-normalization (BN) of the fully connected layer of auxiliary classifier. Such factorization offers significant computational cost savings through sharing of weights between the sub-layers while the BN serves as a good regularizer. The potency of these adaptions can be evaluated by the fact that these architectures have achieved top-5 error rate of 3.46% on the ILSVRC2012's validation data beating the human-level[3] error rate of 5.1%.

The bottleneck features retrieved from the antepenultimate layer of the Inception v3 architecture are used to train a **SVM** and a **GB** classifier as final recognizers (Figure 1). Our idea of replacement of the softmax layer as the final classifier is derived from several image classification works [25, 26] wherein, the use of third-party classifiers have been known to advance the generalization capacity of the original models.

#### 3.2.2 Proposed Method

Our method for cartoon face recognition can be described in two major phases, as depicted in Figure 2a. **Phase I** deals with annotation of the coordinates of 15 facial keypoints over following steps:

---
[3]https://karpathy.github.io/2014/09/02/what-i-learned-from-competing-against-a-convnet-on-imagenet/

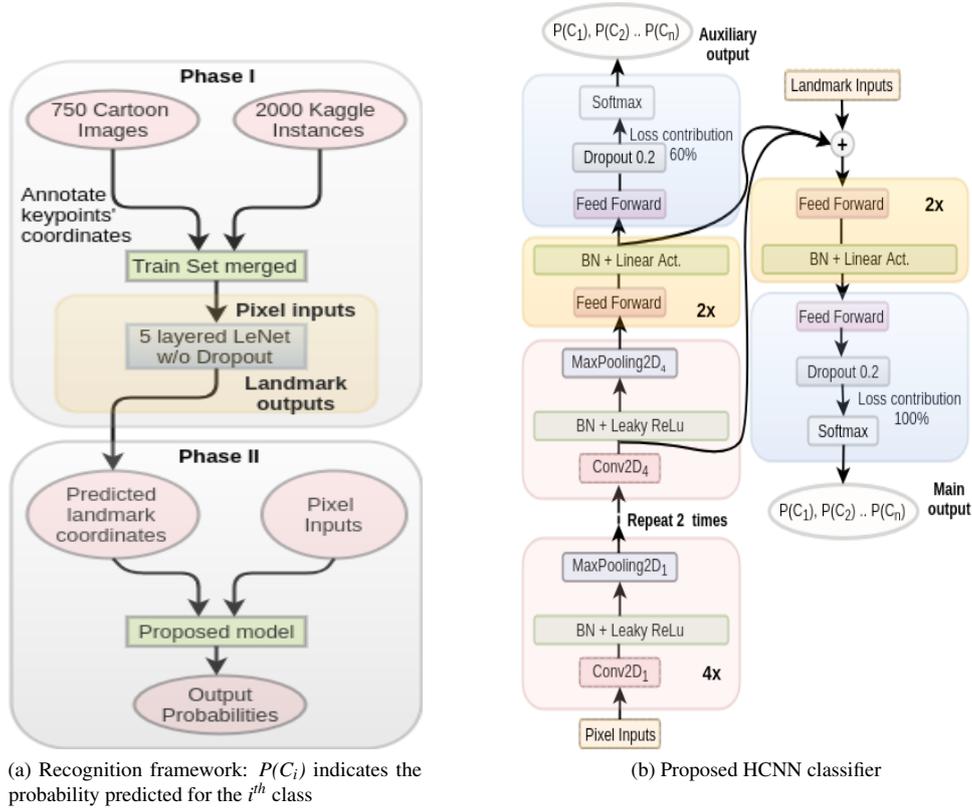

(a) Recognition framework: $P(C_i)$ indicates the probability predicted for the $i^{th}$ class

(b) Proposed HCNN classifier

Figure 2: Schematic diagram of the proposed framework built upon the landmark extraction concept and the HCNN classifier employed within.

1. *Pre-processing:* The cartoon images are first grayscaled and normalized.

2. *Landmark extraction*: The location coordinates of 15 facial landmarks of 750 normalized images belonging to 50 cartoon classes (15 for each) were manually annotated by three undergraduates using a Java swing application. The images were selected irrespective of their facial posture. Any absent or unlocatable keypoint was assigned a null value. The landmarks follow the ordering listed in the Kaggle Facial Keypoints Detection Challenge[4] database:
*left eye center, right eye center, left eye inner corner, left eye outer corner, right eye inner corner, right eye outer corner, left eyebrow inner end, left eyebrow outer end, right eyebrow inner end, right eyebrow outer end, nose tip, mouth left corner, mouth right corner, mouth center top lip, mouth center bottom lip*

   For verification of the annotations, each of the coordinates were cross-checked to be within the valid range[5]. (Database in supplementary materials.)

3. *Landmark detection:* For predicting the landmarks of rest of the cartoon images in the database, we employ the 5-layer LeNet architecture with dropout, as described in Longpre and Sohmshetty [27]. Since, the labelled landmark data for cartoon faces is too small for training the architecture, we further merge it with 2000 real human instances of the Kaggle's database.

---
[4]https://www.kaggle.com/c/facial-keypoints-detection
[5]Valid range for each feature was decided using the mean value for that column.

Given the landmark positions of each image, **phase II** then leverages these for face recognition using a proposed CNN model, which we refer to as the **hybrid CNN** (HCNN) model. The HCNN model, as depicted in Figure 2b, adapts the abstract stack structure of LeNet architectures. The model architecture was chosen through rigorous experiments. The inputs comprise of pixel values of the normalized image and the 30 features forming the location coordinates. As practised in [27], we employ four stacks of alternating convolution and max pooling layers. The filter shapes of the convolutional layers descend from $Conv2D_1$ with (4,4) to $Conv2D_4$ with (1,1) while all the maxpooling layers have a pool shape of (2,2) with non-overlapping strides and without zero padding. The output of the Conv2D stack is fed to two subsequent feed forward layers following which, an additional such layer with softmax activation serves as the auxiliary classifier with a discounted loss of 0.60 added to the total training loss whereas, three more such dense layers process the concatenated inputs to form the main classifier. All the Conv2D layers use Leaky Rectified Linear Unit (ReLu) activations [28] to resolve the problem of dead ReLus seen during initial test runs. All but the final dense layers employ linear activations so that the non-linear outputs of the previous layers do not remain restricted to the positive domain.

We apply BN [29] in between the network layers and their activations. Backed by successive experiments, we drop the dropout layers throughout the inner layers of the network as their combination with BN further degraded the performance of the model (Section 4.5.1). Dropouts are applied only to the feed forward layers right before the final softmax activations considering that there are no subsequent layers incorporating BN [30]. A shortcut connection [31] concatenates the outputs of the final convolutional layer with that of the landmark inputs.

Each convolutional layer is assigned a randomly initialized weight. The weights of all the dense layers are initialized using the Glorot uniform initialization [32].

The main classifier uses the Adam optimization [33] technique with a learning rate of 0.001, $beta_1$ of 0.9, $beta_2$ of 0.999, epsilon of $10^{-8}$ while the auxiliary classifier employs the Stochastic Gradient Descent optimization [34] with a Nesterov momentum of 0.9, a weight decay of 0.0001, and a learning rate starting from 0.2 and being divided by 10 as the error plateaus over 1,000 epochs. Both the classifiers minimize the categorical cross-entropy loss throughout training.

## 4 EXPERIMENTAL SETTINGS

Experiments were carried out from two points of view to investigate the performance of face detection and recognition models independently. Our system consists of x86_64 GNU/Linux with 8G memory using one NVIDIA GeForce 840M with CUDA V8.0.61, Tensorflow-1.4.1, Keras-2.0.6, and OpenCV V3.4.0.

The 2048-D feature vectors extracted from the Inception v3 model are normalized into [0,1] range using min-max normalization. Table 1 describes the best model parameters of the SVM and the GB classifiers determined using the Grid Search algorithm [35] over 10-fold cross validation on 100 input classes.

Table 1: Settings for SVM and GB assisted model parameters

| Classifier | Parameter Setting |
|---|---|
| SVM | Penalty parameter = 50, Kernel = Radial basis function (RBF), Kernel coefficient = $10^{-3}$, Probability estimates = enabled |
| GB | Loss function = 'deviance' for probabilistic outputs, Shrinkage contribution of each tree = 0.08, Maximum depth of the individual regression estimators = 3, No. of boosting stages = 100 |

### 4.1 Datasets

We use the benchmark IIIT-CFW (Cartoon Faces in the Wild) database containing 8,928 annotated images of cartoon faces belonging to 100 global public figu res. The annotations consist of face bounding boxes estimated manually along with some additional attributes such as age group, view, expression, pose etc. for each image. For the character and gender recognition tasks, we carry out 80:20 train:test splits based on class-wise and gender-wise manners respectively. A validation split of 0.1 is further made on the train sets.

Considering the insufficient cartoon face instances in IIIT-CFW database, we use the CASIA WebFace Database[6] for training the MTCNN model. The Haar feature-based Cascade Classifier was trained using 3,000 positive and negative normalized images for 50 stages. Positive samples comprised entirely of the cartoon images while the negative samples were a blend of 750 images of fish, flower, utensils and beverages each, extracted from the ImageNet URL links.

### 4.2 Data Augmentation

The unequal instances of cartoon characters provided in the IIIT-CFW database introduces class-imbalance with the number of images varying from as many as 299 to as few as 11 for different celebrities. A similar problem persists in the case of gender-wise split whereby, the number of male and female faces in the train set amount to 5,242 and 1,896 respectively. It is worth noting that the train sets integrate the real human faces of celebrities included within the database. While we leave the test sets unchanged, we take the following measures to attenuate the effect of biased training of the recognition models due to the imbalance:

1. For the character recognition task, each class is allowed to have a maximum of 800 and a minimum of 600 instances prior to training. The images are over-sampled using a blend of three augmentation steps: horizontal and/or vertical flip of the images followed by horizontal and/or vertical shifts, and rotation of images. The shifts are performed in a range of 30% of the original width and height of the images while the rotation range is within 30 degrees. For training the HCNN model, the coordinate locations of the landmark points are adjusted accordingly.

2. For the gender recognition task, the same oversampling steps are applied to the female faces until the ratio of male:female instances become 1:1.

---
[6]http://www.cbsr.ia.ac.cn/english/CASIA-WebFace-Database.html

### 4.3 Face Detection Results

We use the bounding box information provided in the annotations as ground truth values. The implementation of Haar features-based detector relies upon OpenCV while that of HOG features-based detector is based upon the dlib library.

#### 4.3.1 Evaluation Measures

We count on three evaluation measures for the detection systems: True Positive Rate (TPR) relates having detected a right facial region, False Positive Rate (FPR) relates having detected wrong regions in addition to a right face region, and False Negative Rate (FNR) relates not having detected a right face region. Table 2 shows a comparison of the results obtained by the employed models. For a fair comparison with the state-of-the-art (Takayama *et al.*) [14], we compute the rates by averaging over an input of 500 color images (5 images for each class) selected from the database.

The jaw contour and facial symmetry based method of Takayama *et al.* reports a TPR of 74.2% which is the highest reported yet to the best of our knowledge. Along with outperforming it in terms of TPR, the MTCNN based method also delivers lesser FPR and FNR score. While the Haar-features result the poorest in correctly detecting the right facial regions, it outperforms HOG in terms of being confused for detecting the non-facial regions. Also, it is worth noting that the results of the OpenCV classifier mentioned in Takayama *et al.* vary greatly from ours. While they do not mention the training strategy, a plausible explanation could be the classifier being pre-trained on real human faces instead of cartoons.

As depicted in Figure 3, each of the classifiers possess their own failure cases. Moreover, there are multiple such instances where the classifiers detect no facial regions at all. These act as bottlenecks in using the detected regions for the face recognition experiments. Hence, we use the original bounding box annotations for extracting the facial regions prior to face recognition.

### 4.4 Face Recognition Results

We demonstrate the performance of both the recognition models on the task of personality and gender recognition of cartoon faces alongside evaluating the performance of the 5-layer LeNet architecture [27] for landmark extraction. The inputs are normalized after extracting the face bounding box.

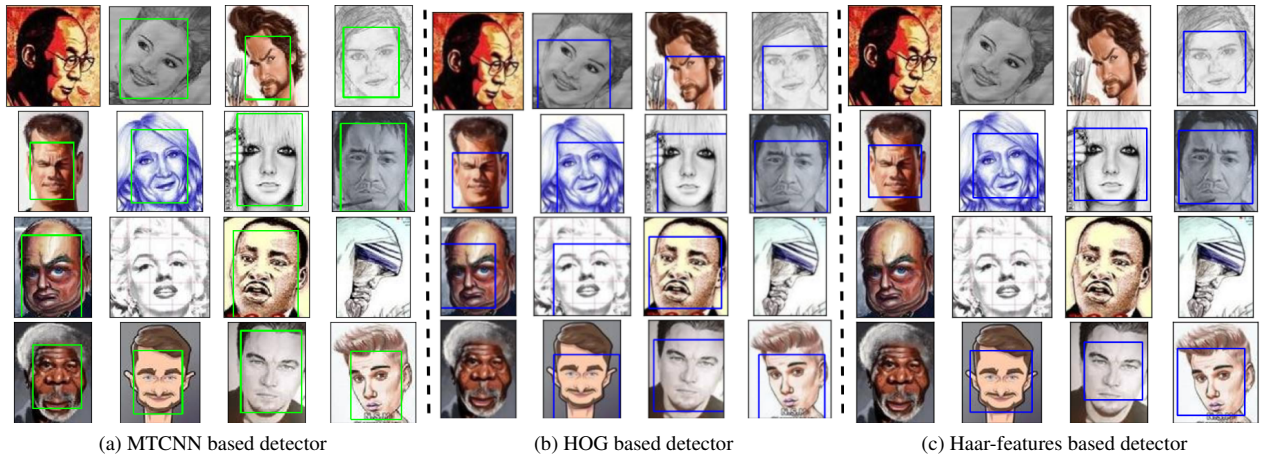

(a) MTCNN based detector    (b) HOG based detector    (c) Haar-features based detector

Figure 3: Figure showing outputs of different face detection methods.

Table 2: Face detection scores of various methods

| Models | Randomly chosen | | | Frontal | | |
|---|---|---|---|---|---|---|
| | TPR | FPR | FNR | TPR | FPR | FNR |
| MTCNN | 78.17% | 12.81% | 9.02% | 83.67% | 4.90% | 11.43% |
| HOG features | 70.51% | 17.33% | 12.16% | 77.32% | 11.80% | 10.88% |
| Haar-features | 57.24% | 8.39% | 34.37% | 69.44% | 3.05% | 27.51% |
| Takayama et al. | - | - | - | 74.2% | 14.0% | 11.8% |

#### 4.4.1 Evaluation Measures

Precision, recall and F-measure averaged over all the classes are used as primary quality assessment metrics. Additionally, we evaluate the accuracy and top-5 error rate of each model. The performance of the landmark extraction system is evaluated in terms of the Root Mean Squared Error (RMSE) measured in between the actual and the predicted landmarks.

#### 4.4.2 Landmark Extraction system

We perform a 80:20 split of the merged database (Section 3.2.2) to obtain the train and validation sets. Table 3 shows a comparison of validation RMSEs of the LeNet based model [27] on three different train sets: (i) the standard train set consisting a blend of cartoon and human faces (ii) only the 750 cartoon faces (iii) only the human faces (as presented in [27]). The validation RMSE degrades by ~3.4 times on the blended train set, and worsens further when (ii) is used for training. While the mapping of landmark features on cartoons could require greater degrees of non-linear approximations than humanly faces [36], the degradation can be attributed to the fewer size of annotated cartoon samples as well. Further increase of human instances in the train set worsen the scores as the predictions gain more of human-like patterns.

Table 3: Validation RMSE comparison of 5-layer LeNet architecture

| Input | Validation RMSE |
|---|---|
| Cartoon + Real Human faces | 8.85 |
| Only cartoons | 19.01 |
| Longpre and Sohmshetty | **2.63** |

Figure 4 depicts such instances where the model fails miserably in locating the keypoint co-ordinates using (i). More closer analysis imply that the images with low intensity boundaries, anomalous organ shapes, missing facial regions and extremely exaggerated features in the database serve majorly towards the performance degradations. Images with lower resolutions, by contrast, do not result in any significant decline unless the aforesaid characteristics are present.

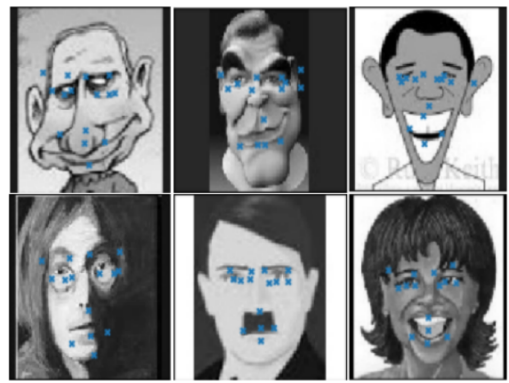

Figure 4: Erroneously predicted landmark locations on various faces

### 4.5 Character Recognition Results

Table 4 presents the accuracy and top-5 error rates of SVM and GB classifiers on the Inception v3 features over 20, 50 and 100 classes, and contrasts them with the results of using the original softmax classifier instead. While the use of SVM improves the top-5 error by 6.909% over the softmax layer, the GB assisted ensemble classifier consistently lags in terms of both metrics. Further, the lag in performance of the GB classifier with respect to SVM increases as the number of classes increase. This lag is in the favor of the conjecture that boosting methods, because of their high correlation with the noise present in the data set, can show a zero gain or even a decrease in performance from a single classifier [37]. We further observe such decrement becoming more evident as the diversity of the training instances flourish.

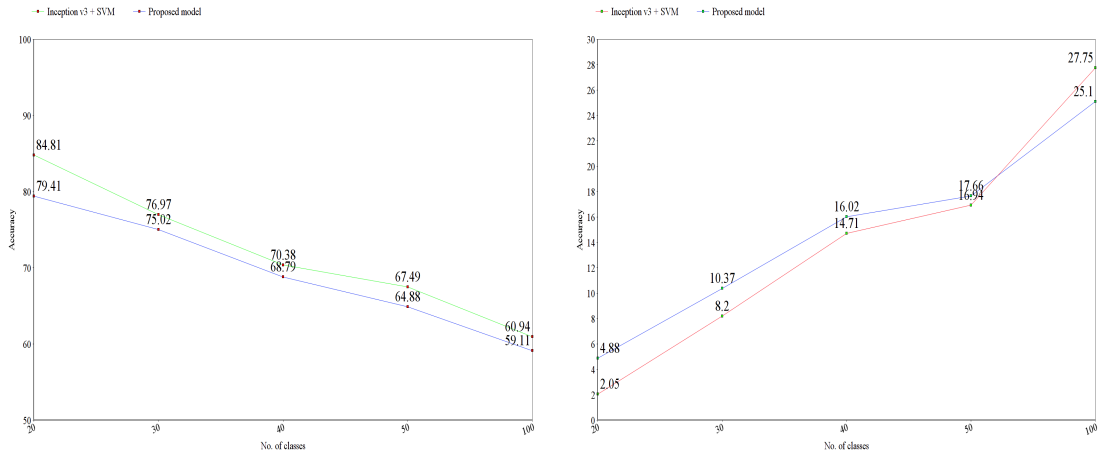

(a) Accuracy comparison      (b) Top-5 error rate comparison

Figure 5: Accuracy and Top-5 error rate comparison between the Inception+SVM model and the proposed model over 20, 30, 40, 50 and 100 cartoon classes

Table 4: Accuracy and Top-5 error rate comparison of Inception-based frameworks

| Models | Accuracy | | | Top-5 error | | |
|---|---|---|---|---|---|---|
| | 20 | 50 | 100 | 20 | 50 | 100 |
| Inception v3 | 79.43% | 63.98% | 56.61% | 5.114% | 22.307% | 34.659% |
| Inception v3+SVM | 84.81% | 67.49% | 60.94% | 2.051% | 16.944% | 27.750% |
| Inception v3+GB | 66.44% | 48.49% | 34.51% | 12.528% | 37.871% | 63.798% |

We do not emphasize the results of incorporating the landmark features into the SVM classifier as doing so, resulted in the drop of error rate to approx. 43% on 100 classes, which is worse than the original softmax classifier.

Table 5 compares the performance of both of our recognition models with the current state-of-the-art [14]. In contrast to [14] reporting their results on 300 cartoon images belonging to 150 different characters, our evaluations leverage 1786 images belonging to 100 characters. Our systems nonetheless outperform theirs (Section 2) with a high margin. It is discernible that although the predictions retrieved by Inception+SVM model contain higher fraction of relevant cartoon classes (high precision), the greater recall score of the HCNN model suggests its higher resistance to mistaking the relevant cartoon classes for irrelevant ones.

Table 5: Performance comparison of recognition models

| Model | Precision | Recall | F-Measure |
|---|---|---|---|
| Proposed Model | 0.622 | **0.680** | 0.649 |
| Inception v3+SVM | **0.682** | 0.659 | **0.670** |
| Takayama et al. | 0.476 | 0.563 | 0.516 |

Table 6 shows the performance comparison of the main and auxiliary classifiers of the HCNN model over 100 classes. The inclusion of the keypoint features impart respective gains of 2.66% and 5.87% to the accuracy and top-5 error rate of the main classifier over the auxiliary classifier. Our experiments show that on dropping of the BN and skip connections, the gain in accuracy decreases to 2.51%, while the error rate witnesses an increment to 6.08%. Overall, we notice that the presence of auxiliary classifier imparts greater stability to the model by increasing the number of training epochs before convergence, and keeping the validation error from larger fluctuations afterwards.

Table 6: Results of the main and auxiliary classifiers of the HCNN model

| Classifier type | Accuracy | Top-5 error |
|---|---|---|
| Main | **59.11%** | **25.10%** |
| Auxiliary | 56.45% | 30.97% |

Figure 5 juxtaposes the accuracy and the top-5 error rates of the models on varying number of classes. The graph in fig. 5a suggests a rather striking trend: though the accuracy of the HCNN model remains significantly lower than that of Inception v3+SVM for lesser number of classes (i.e., 5.4% for 20 classes), their differences diminish to as low as 1.83%, as the number of classes approach to 100. A similar trend can be observed from fig. 5b wherein, the top-5 error rate of HCNN lags noticeably beyond Inception v3+SVM for lesser number of classes (20-40), becomes almost identical for 50 classes and eventually outperforms the latter for 100 classes. These trends, apart from being successively procurable from one another, show that the Inception v3+SVM offers lesser stability as the number of classes vary and as the diversity of the input instances increase. We suggest two tenable explanations for such poor stability of the model. Firstly, the RBF kernel employed in the SVM assumes that the optimal decision boundary remains smooth in all the instances. However, with greater number of classes, the violations of this assumption elevate leading to rise in the entropy captured by the model's hyperparameters due to the diverse dispersion of outliers than earlier. Secondly, for lesser number of classes, the HCNN model with parameters much greater than SVM might have overfitted due to smaller train set size.

### 4.5.1 Effects of skip connection and BN on HCNN:

We experiment with different combinations of skip connections applied to the outputs of each Conv2D and BN layer, one at a time while preserving the original destination layer. We notice that the validation loss increases as the connection is placed right after the BN layers, than when they are applied after the Conv2D layers. The validation loss and the training epochs before convergence increase as the connections are made in presence of BN along with dropouts. The skip connection plays a rather important role in stabilizing the training epochs for convergence, as its elimination increases the number of epochs to 103 (and the top-5 error rate to 36%) while on its presence, these reduce drastically to 47.

### 4.6 Gender Recognition Results

Table 7 depicts the performance of the recognition models on the task of gender classification of cartoon faces. The Inception v3+SVM model clearly outperforms the HCNN on all the three metrics. The lower recall scores for both the models suggest that the mistaken instances of genders for their counterparts are comparatively higher. The scores of the Inception v3+SVM model on the binary classification task further strengthens the argument mentioned in Section 4.5 explaining for its poor stability as the number of classes grow in the character recognition task. At the time of writing, no previous work in the literature talks of such experiment on cartoon faces and thus, we believe that the scores hold a state-of-the-art.

Table 7: Performance comparison of models for gender recognition

| Model | Precision | Recall | F-Measure |
|---|---|---|---|
| Proposed Model | 0.904 | 0.827 | 0.864 |
| Inception+SVM | **0.927** | **0.894** | **0.910** |

## 5 CONCLUSION

Towards the end goal of improving cartoon face detection and recognition systems with the latest advancements in deep learning frameworks, we present the following contributions:

- For the face detection task, we show that the MTCNN framework outperforms the state-of-the-art [14] in terms of TPR, FPR and FNR. We further confirm that the model performs the best when presented with frontal faces.

- For the face recognition task, firstly, we show that a combination of Inception v3 as feature extractor followed by SVM as feature recognizer achieves a benchmark F-score of 0.670. Secondly, we propose a LeNet inspired CNN framework that helps us achieve a more stable top-5 error rate than the former as the number of cartoon classes elevate.

We further suggest intuitions behind the pitfalls of the SVM and GB based classifiers, and depict the differences of employing the LeNet architecture for extracting the landmarks of cartoon faces and real human faces. Our experiments demonstrate that the inclusion of the facial keypoint locations can help improve the top-5 error rate of the proposed recognition model by 5.87%. The annotated facial keypoints are thus made publically available in hope to aid further researches related to the field. Lastly, we show that the Inception+SVM establishes a state-of-the-art F-measure of 0.927 when employed to the task of gender recognition of cartoon faces.

## REFERENCES

[1] K. Zhang, Z. Zhang, Z. Li, and Y. Qiao, "Joint face detection and alignment using multitask cascaded convolutional networks," *IEEE Signal Processing Letters*, vol. 23, pp. 1499–1503, 2016.

[2] C. Szegedy, V. Vanhoucke, S. Ioffe, J. Shlens, and Z. Wojna, "Rethinking the inception architecture for computer vision," in *Proceedings of the IEEE Conference on Computer Vision and Pattern Recognition*, 2016, pp. 2818–2826.

[3] A. Mishra, S. N. Rai, A. Mishra, and C. Jawahar, "Iiit-cfw: A benchmark database of cartoon faces in the wild," in *European Conference on Computer Vision*. Springer, 2016, pp. 35–47.

[4] K. Habib and T. Soliman, "Cartoons effect in changing children mental response and behavior," *Open Journal of Social Sciences*, vol. 3, no. 09, p. 248, 2015.

[5] E. Ibili and S. Sahin, "The use of cartoons in elementary classrooms: An analysis of teachers' behavioral intention in terms of gender," *Educational Research and Reviews*, vol. 11, no. 8, p. 508, 2016.

[6] P. A. Viola and M. J. Jones, "Robust real-time face detection," *International Journal of Computer Vision*, vol. 57, pp. 137–154, 2001.

[7] H. A. Rowley, S. Baluja, and T. Kanade, "Neural network-based face detection," in *CVPR*, 1996.

[8] R.-L. Hsu, M. Abdel-Mottaleb, and A. K. Jain, "Face detection in color images," *IEEE Trans. Pattern Anal. Mach. Intell.*, vol. 24, pp. 696–706, 2001.

[9] W. Zhao, A. Krishnaswamy, R. Chellappa, D. L. Swets, and J. Weng, "Discriminant analysis of principal components for face recognition," in *Face Recognition*. Springer, 1998, pp. 73–85.

[10] T. Ahonen, A. Hadid, and M. Pietikainen, "Face description with local binary patterns: Application to face recognition," *IEEE transactions on pattern analysis and machine intelligence*, vol. 28, no. 12, pp. 2037–2041, 2006.

[11] J. Wright, A. Y. Yang, A. Ganesh, S. S. Sastry, and Y. Ma, "Robust face recognition via sparse representation," *IEEE transactions on pattern analysis and machine intelligence*, vol. 31, no. 2, pp. 210–227, 2009.

[12] R. K. McConnell, "Method of and apparatus for pattern recognition," Jan. 28 1986, uS Patent 4,567,610.

[13] P. A. Viola and M. J. Jones, "Rapid object detection using a boosted cascade of simple features," in *CVPR*, 2001.

[14] K. Takayama, H. Johan, and T. Nishita, "Face detection and face recognition of cartoon characters using feature extraction," in *Image, Electronics and Visual Computing Workshop*, 2012, p. 48.

[15] B. E. Boser, I. M. Guyon, and V. N. Vapnik, "A training algorithm for optimal margin classifiers," in *Proceedings of the fifth annual workshop on Computational learning theory*. ACM, 1992, pp. 144–152.

[16] J. H. Friedman, "Greedy function approximation: a gradient boosting machine," *Annals of statistics*, pp. 1189–1232, 2001.

[17] R. Glasberg, A. Samour, K. Elazouzi, and T. Sikora, "Cartoon-recognition using video & audio descriptors," in *Signal Processing Conference, 2005 13th European*. IEEE, 2005, pp. 1–4.

[18] R. Glasberg, S. Schmiedeke, M. Mocigemba, and T. Sikora, "New real-time approaches for video-genre-classification using high-level descriptors and a set of classifiers," in *Semantic Computing, 2008 IEEE International Conference on*. IEEE, 2008, pp. 120–127.

[19] B. Ionescu, C. Vertan, P. Lambert, and A. Benoit, "A color-action perceptual approach to the classification of animated movies," in *Proceedings of the 1st ACM International Conference on Multimedia Retrieval*. ACM, 2011, p. 10.

[20] N.-V. Nguyen, C. Rigaud, and J.-C. Burie, "Comic characters detection using deep learning," *2017 14th IAPR International Conference on Document Analysis and Recognition (ICDAR)*, vol. 03, pp. 41–46, 2017.

[21] J. Redmon and A. Farhadi, "Yolo9000: better, faster, stronger," *arXiv preprint*, vol. 1612, 2016.

[22] W.-T. Chu and W. Li, "Manga facenet: Face detection in manga based on deep neural network," in *ICMR*, 2017.

[23] H. Li, Z. Lin, X. Shen, J. Brandt, and G. Hua, "A convolutional neural network cascade for face detection," in *Proceedings of the IEEE Conference on Computer Vision and Pattern Recognition*, 2015, pp. 5325–5334.

[24] S. Ioffe and C. Szegedy, "Batch normalization: Accelerating deep network training by reducing internal covariate shift," in *International conference on machine learning*, 2015, pp. 448–456.

[25] X.-X. Niu and C. Y. Suen, "A novel hybrid cnn–svm classifier for recognizing handwritten digits," *Pattern Recognition*, vol. 45, no. 4, pp. 1318–1325, 2012.

[26] D.-X. Xue, R. Zhang, H. Feng, and Y.-L. Wang, "Cnn-svm for microvascular morphological type recognition with data augmentation," *Journal of medical and biological engineering*, vol. 36, no. 6, pp. 755–764, 2016.

[27] S. Longpre and A. Sohmshetty, "Facial keypoint detection," 2016.

[28] A. L. Maas, "Rectifier nonlinearities improve neural network acoustic models," 2013.

[29] S. Ioffe and C. Szegedy, "Batch normalization: Accelerating deep network training by reducing internal covariate shift," in *ICML*, 2015.

[30] X. Li, S. Chen, X. Hu, and J. Yang, "Understanding the disharmony between dropout and batch normalization by variance shift," *CoRR*, vol. abs/1801.05134, 2018.

[31] K. He, X. Zhang, S. Ren, and J. Sun, "Deep residual learning for image recognition," in *Proceedings of the IEEE conference on computer vision and pattern recognition*, 2016, pp. 770–778.

[32] X. Glorot and Y. Bengio, "Understanding the difficulty of training deep feedforward neural networks," in *AISTATS*, 2010.

[33] D. P. Kingma and J. Ba, "Adam: A method for stochastic optimization," *CoRR*, vol. abs/1412.6980, 2014.

[34] L. Bottou, "Large-scale machine learning with stochastic gradient descent," in *Proceedings of COMPSTAT'2010*. Springer, 2010, pp. 177–186.

[35] J. Bergstra and Y. Bengio, "Random search for hyper-parameter optimization," *Journal of Machine Learning Research*, vol. 13, no. Feb, pp. 281–305, 2012.

[36] M. Arai, "Feature extraction methods for cartoon character recognition," in *Image and Signal Processing (CISP), 2012 5th International Congress on*. IEEE, 2012, pp. 445–448.

[37] D. W. Opitz and R. Maclin, "Popular ensemble methods: An empirical study," *J. Artif. Intell. Res.(JAIR)*, vol. 11, pp. 169–198, 1999.